\documentclass[11pt]{article}

\usepackage[final]{acl}

\usepackage{times}
\usepackage{latexsym}

\usepackage[T1]{fontenc}

\usepackage[utf8]{inputenc}

\usepackage{microtype}

\usepackage{inconsolata}

\usepackage{graphicx}

\usepackage{amsmath}
\usepackage{subcaption}
\usepackage{booktabs}
\usepackage{listings}
\usepackage[most]{tcolorbox}
\usepackage{multirow}
\usepackage{multicol}
\usepackage{amssymb}
\usepackage{siunitx}
\usepackage{makecell}
\title{Decoupling Turn-Taking from Semantics: A Decoupled Data Approach for Finite-State-Machine-Based Full-Duplex Dialogue}

\author{Yihang Li, Chenhui Chu \\
  Kyoto University \\
  \texttt{liyh@nlp.ist.i.kyoto-u.ac.jp}, \texttt{chu@i.kyoto-u.ac.jp}
}

\begin{document}
\maketitle

\begin{abstract}
The Neural Finite State Machine (NFSM) framework offers a pragmatic path to full-duplex dialogue by serializing turn-taking control and response generation onto a single causal tape under the standard next-token prediction objective, thereby preserving semantic prowess at a low fine-tuning cost. However, its reliance on synthetic text data fundamentally limits turn-taking naturalness, as Large Language Models (LLMs) cannot faithfully simulate the fine-grained acoustic temporal dynamics of real human dialogues. In this work, we propose a decoupled data approach that learns turn-taking from real Human-Human (HH) spoken dialogues while shaping semantic behavior through configurable Human-Agent (HA) text dialogues. To operationalize this approach, we introduce a rule-based event-guided data transformation method that serializes HH spoken dialogues into FSM tapes by classifying turn-taking events and applying deterministic mapping rules, enabling scalable supervision without LLM-generated annotations. We further propose a Source-Aware Calibrated (SAC) Loss that jointly calibrates the long-tailed distribution of state transition tokens and channels each data source toward the capability it best supervises. Experiments show that our approach substantially improves turn-taking proficiency while recovering the foundation LLM's semantic capability. Our code and model are available at \url{https://github.com/Liyht/def-fsm}.
\end{abstract}

\section{Introduction}

Recent advances in Large Language Models (LLMs) have greatly enhanced the semantic capabilities of voice assistants \cite{ji2024wavchatsurveyspokendialogue}. However, their turn-taking naturalness remains constrained by rigid half-duplex turn-taking, in which each party must wait for the other to finish before speaking. To bridge the gap toward truly human-like interaction, the field is moving toward full-duplex dialogue \cite{chen2025turntakingsynchronousdialoguesurvey}, where an agent can listen and speak simultaneously and thereby handle interruptions, acoustic backchannels, and dynamic turn-taking in real time.

\begin{figure}[t]
  \centering
  \includegraphics[width=1.0\linewidth]{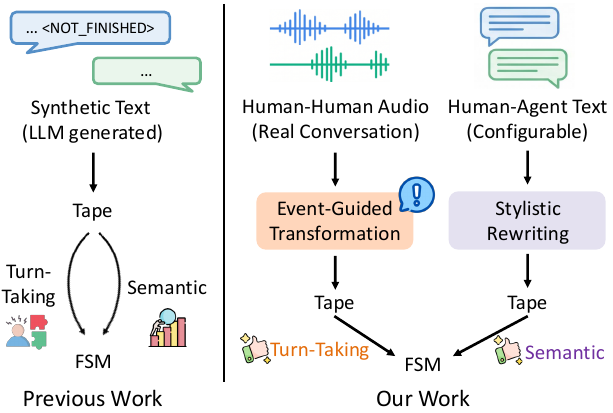}
  \caption{Comparison between the coupled and decoupled data approaches.}
  \vspace{-3mm}
  \label{fig:approach}
\end{figure}

Current research on full-duplex dialogue falls into two primary paradigms: end-to-end and modular.
While the end-to-end paradigm \cite{nguyen-etal-2023-generative, moshi, veluri-etal-2024-beyond, omniflatten, lee-etal-2025-behavior} excels in fine-grained turn-taking control and low latency by directly understanding and generating speech representations, enforcing strict cross-modal alignment often causes significant degradation in the underlying LLM's semantic capabilities \cite{xiang-etal-2025-understanding, chen2024voicebenchbenchmarkingllmbasedvoice}.
The modular paradigm instead equips an LLM with auxiliary state management via external controllers \cite{flexduo}, classification heads \cite{freezeomni, minmo}, or in-vocabulary FSM control tokens \cite{nfsm}.
Among these approaches, the NFSM \cite{nfsm} occupies a uniquely minimal position: rather than introducing any auxiliary module, it serializes both state transitions and response content onto a single causal ``tape,'' handling full-duplex control entirely under the standard next-token prediction objective with no added parameters. Moreover, it enables the LLM to explicitly perceive and condition on the full history of prior turn-taking dynamics when making subsequent decisions. However, this minimalism places the full burden of turn-taking competence on the tape's textual content—a burden that the original NFSM's data approach cannot adequately meet.

Specifically, as illustrated in Figure \ref{fig:approach} (left), NFSM constructs training tapes from LLM-generated transcripts, attempting to distill complex turn-taking dynamics and rich semantics simultaneously. However, as foundation LLMs are predominantly pre-trained on structured non-overlapping text, they lack the inherent ability to simulate the fine-grained acoustic and temporal dynamics of real human interactions. Worse still, jointly controlling realistic turn-taking and rich semantics within a single generated dialogue compounds this difficulty, making such synthetic data fundamentally unreliable as supervision.

To address this challenge, we advocate a decoupled data approach, as shown in Figure \ref{fig:approach} (right), that aligns each capability with its most suitable data source: fine-grained turn-taking dynamics are acquired from real HH spoken dialogue corpora, while readily available HA text dialogues serve as a flexible lever for shaping semantic behavior. The central challenge in operationalizing this approach lies in transforming real HH spoken dialogues into NFSM's serialized tape as ground truth, in which state transition tokens must be inserted to faithfully reflect complex acoustic turn-taking dynamics. To this end, we propose an \emph{event-guided data transformation} strategy that segments and classifies the dialogue timeline into discrete turn-taking events and subsequently applies deterministic mapping rules to serialize each segment into the tape. Our core insight is that state transition token insertion acts as an approximation of the turn-taking dynamics, while turn-taking dynamics equate to the temporal sequence of turn-taking events. Crucially, the entire pipeline is fully rule-based and free of LLM annotations, making it readily scalable to arbitrarily large HH corpora. On the HA side, the corpus can be freely substituted to target specific semantic objectives. In this work, we adopt a generic HA corpus to preserve the LLM's intrinsic semantic capabilities, leaving capability-specific adaptation as a natural extension.

While this decoupled data approach provides clean supervision signals, jointly training on the resulting heterogeneous tapes introduces two challenges. First, state transition tokens exhibit an extremely long-tailed distribution, in which rare yet operationally critical state-switch tokens are easily marginalized under standard cross-entropy training. Second, the two data sources carry asymmetric supervisory value: HH tapes are authoritative for turn-taking but noisy for semantics, whereas HA tapes carry reliable semantics but trivial turn-taking structure. To address both issues within a single objective, we introduce the Source-Aware Calibrated (SAC) Loss, which combines logit adjustment over the restricted vocabulary of state transition tokens with a source-aware weighting scheme that channels each data source toward the capability it best supervises.

Experiments under strict token-volume constraints on HH spoken dialogue corpora and the VoiceBench dataset show that our model achieves substantially better turn-taking proficiency and semantic capability than the conventional synthetic-data baseline, with SAC further mitigating the severe class imbalance. When scaled to a larger training corpus constructed by proportionally upsampling each data source to preserve the mixture ratio, our approach successfully recovers the intrinsic semantic capability of the foundation LLM.

In summary, our contributions are threefold:
\begin{itemize}
    \item We propose a decoupled data approach for FSM-based full-duplex dialogue, in which turn-taking is learned from real HH spoken dialogues and semantic capability is independently shaped through configurable HA data.
    \item We design a fully rule-based event-guided transformation that serializes complex acoustic interactions into causal FSM tapes, enabling scalable construction of real HH dialogue data.
    \item We introduce the SAC Loss, which calibrates the long-tailed distribution of state transition tokens and specializes optimization across data sources.
\end{itemize}

\section{Related Work}

\textbf{Turn-Taking Event Classification.}
Turn-taking event classification is a labeling scheme that constructs ground-truth event labels along the dialogue timeline, supporting downstream tasks such as event prediction and analysis \cite{technologies13120591, castillo-lopez-etal-2025-survey, Heldner2010PausesGA, arora2025talking, threlkeld-etal-2022-using}. Existing schemes can be organized by the temporal unit on which events are defined. Chunk-based methods segment the timeline into equal-length frames and assign per-frame labels \cite{Heldner2010PausesGA, Ekstedt2022VoiceAP, arora2025talking}, offering fine temporal resolution suited to frame-based neural architectures. Inter-Pausal Unit (IPU)-based methods instead define events at IPU boundaries delimited by silence thresholds \cite{ekstedt-skantze-2020-turngpt, wang2024turn}, yielding linguistically coherent units that align cleanly with textual chunk boundaries. We adopt the IPU-based formulation, as this property makes rule-based serialization onto the FSM tape straightforward.

\textbf{Data Strategies for Full-Duplex Supervision.}
Full-duplex supervision data are constructed along two distinct paths.
The first relies on \emph{synthetic generation}: an LLM is prompted to produce dialogues annotated with designated turn-taking control tokens \cite{nfsm}, or dialogues are assembled through rule-based procedures such as inserting user interruptions at predefined positions \cite{miniomni2}.
The fundamental difficulty is that LLMs are trained predominantly on structured non-overlapping text and therefore lack the capacity to simulate the fine-grained turn-taking dynamics of real human conversation.
The second path draws on \emph{real spoken dialogue corpora}.
End-to-end models leverage such corpora through implicit supervision, directly learning from raw interleaved audio tokens without explicit state labels \cite{nguyen-etal-2023-generative, moshi}.
Modular approaches instead apply voice activity detection and heuristic rules to derive state labels from real audio, training a dedicated state classifier on fixed acoustic chunks \cite{flexduo, minmo}.
Our setting poses a qualitatively different challenge: rather than classifying states over fixed acoustic chunks, we must insert discrete state transition tokens into an interleaved text token stream, which is a form of supervision that cannot be derived from voice activity alone and demands explicit turn-taking event identification.

\begin{figure*}[t]
  \centering
  \includegraphics[width=0.97\linewidth]{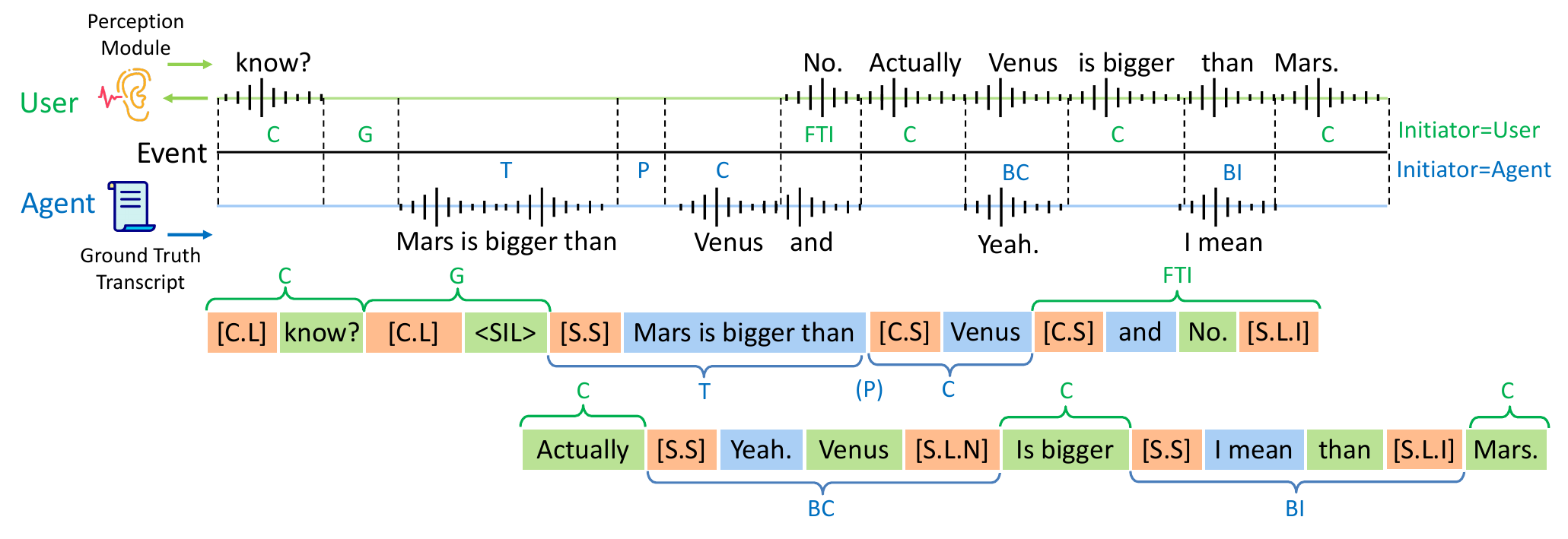}
  \vspace{-2mm}
  \caption{Example of turn-taking event classification and event-guided tape serialization. Green, blue, and orange blocks denote user content, agent content, and state transition tokens (shown in abbreviated form, e.g., \texttt{[C.L]} for \texttt{[C.LISTEN]}), respectively. The \texttt{<user>} role-prefix token preceding each user block is omitted for brevity.}
  \label{fig:transformation}
\end{figure*}

\section{Data Transformation}
Building on the NFSM framework, we serialize two complementary corpora into a single causal FSM tape format: real HH spoken dialogues for fine-grained turn-taking dynamics, and HA text dialogues adapted to a spoken style for preserving semantic capability. We first briefly review NFSM, then describe our transformation pipelines for both corpora.

\subsection{Preliminary: The NFSM Framework}
NFSM \cite{nfsm} casts dialogue turn-taking as an explicit state transition problem managed directly by an LLM operating a two-state machine consisting of a \verb|SPEAK| state and a \verb|LISTEN| state. The transitions are realized through vocabulary tokens: \verb|[C.SPEAK]| and \verb|[C.LISTEN]| for continuing the current state, and \verb|[S.SPEAK]| and \verb|[S.LISTEN]| for switching to the opposite state. Both state transitions and response generation are serialized onto a single causal tape under the standard next-token prediction paradigm. Crucially, each autoregressive decoding step on this tape is triggered by three types of events ranked in decreasing priority: the generation of speech-driving tokens, the arrival of new perception chunks, or the completion of motor processing. This unified sequence asynchronously synchronizes the perception and motor modules, allowing a standard text-based LLM to control real-time full-duplex interaction without architectural modifications.

\subsection{Event-Guided Human-Human Data Transformation}
Building upon our decoupled data approach, the primary objective of HH data transformation is to construct a rigorous ground-truth tape that strictly mirrors human turn-taking behavior. Our core insight is that state transition token insertion acts as an approximation of the turn-taking dynamics, while the turn-taking dynamics is equivalent to the temporal sequence of turn-taking events. 
Consequently, we propose an event-guided transformation strategy. Our conversion strategy is guided by two principles: (1) categorizing the timeline into discrete turn-taking events to direct token insertion, and (2) aligning with the actual inference-time conditions the FSM will encounter. 

\textbf{Asymmetric Preprocessing.} The two channels of each HH dialogue are designated as the user channel and the agent channel, and processed asymmetrically to mirror NFSM's actual inference-time conditions as shown in Figure \ref{fig:transformation}. The user channel is transcribed by the same perception module deployed in the FSM, producing fine-grained timestamped text chunks together with silence tokens of fixed duration, thereby exposing the LLM to realistic ASR granularities and recognition error patterns. The agent channel, in contrast, uses fine-grained timestamped ground-truth transcripts, ensuring the LLM learns to generate clean text rather than imitate ASR artifacts. Chunks are kept as fine-grained as possible—ideally word-level—to enable precise event segmentation downstream. Additional preprocessing details are provided in Appendix~\ref{sec:appendix_preprocess}.

\begin{table*}
  \centering
  \small
    \begin{tabular}{@{}lll@{}}
    \toprule
    Event & Initiator = User & Initiator = Agent \\ \midrule
    T & \verb|[S.LISTEN.N]| <User> & \verb|[S.SPEAK]| <Agent> \\
    C & \verb|[C.LISTEN]| <User> & \verb|[C.SPEAK]| <Agent> \\
    P & \verb|[C.LISTEN]| <SIL> & - \\
    G & \verb|[C.LISTEN]| <SIL> & \verb|[S.LISTEN.N]| <SIL> \\
    BC & \verb|[C.SPEAK]| <Agent> <User> \verb|[C.SPEAK]| & \verb|[S.SPEAK]| <Agent> <User> \verb|[S.LISTEN.N]| \\
    FTI & \verb|[C.SPEAK]| <Agent> <User> \verb|[S.LISTEN.I]| & \verb|[S.SPEAK]| <Agent> <User> \verb|[C.SPEAK]| \\
    BI & \verb|[C.SPEAK]| <Agent> <User> \verb|[C.SPEAK]| & \verb|[S.SPEAK]| <Agent> <User> \verb|[S.LISTEN.I]| \\ \bottomrule
    \end{tabular}
  \caption{Mapping rules for event-guided tape serialization across 14 combinations of turn-taking events and initiators. <User> and <Agent> denote textual content from the user and agent channels; <SIL> represents the silence token.}
  \label{tab:Serialization_rule}
  \vspace{-2mm}
\end{table*}

\textbf{Turn-Taking Event Classification.} This stage partitions the continuous dialogue timeline and assigns each segment to one of seven turn-taking event types, as illustrated in Figure \ref{fig:transformation} (top). We first merge adjacent chunks within each channel into IPUs using a predefined pause threshold; the union of IPU boundaries across both channels then yields a discrete timeline. Following \citet{arora2025talking}, we classify segments as:
\begin{itemize}
    \item \textbf{Single-party speech.} \textit{Turn Change (T)} if the floor holder differs from the previous segment; \textit{Continuation (C)} otherwise.
    \item \textbf{Silence.} \textit{Pause (P)} within a single speaker's turn; \textit{Gap (G)} between alternating turns.
    \item \textbf{Overlapping speech.} \textit{Backchannel (BC)} if the utterance matches a predefined lexical list and falls below a duration threshold \cite{ekstedt-skantze-2020-turngpt, wang2024turn, arora2025talking}; \textit{Floor-Taking Interruption (FTI)} if the interrupter successfully takes the floor; or \textit{Butting-in (BI)} if the attempt fails.
\end{itemize}

\textbf{Event-Guided Tape Serialization.} This stage constructs the serialized tape representation for each classified segment. First, we split \verb|[S.LISTEN]| into \verb|[S.LISTEN.N]| (Natural) and \verb|[S.LISTEN.I]| (Interrupt), granting the LLM finer-grained control over the motor module to explicitly signal whether the ongoing TTS should complete naturally or be immediately truncated. Next, we assign the text of each fine-grained chunk to its corresponding segment based on start timestamp. To formulate the mapping scheme, we define the \emph{initiator} of a segment as the participant whose action triggers the transition from the preceding segment (specifically, the preceding floor holder for gaps). Combining the 7 event types with these 2 initiator types yields 14 deterministic mapping rules (Table \ref{tab:Serialization_rule}; see Figure \ref{fig:transformation} bottom for an example). 
Finally, four operational heuristics govern the compilation of these rules into a continuous tape: (1) the governing state transition token immediately follows the first chunk of a segment, while subsequent chunks maintain the state via continuation tokens; (2) during overlaps, agent chunks precede user ASR chunks, reflecting the causal latency that an agent generates text prior to audio synthesis, whereas the perception module emits text post-speech; (3) silence tokens are preserved exclusively within \verb|LISTEN| intervals; and (4) consecutive state transition tokens at segment boundaries are collapsed into a single token via the rules in Appendix~\ref{sec:appendix_boundary}.

\textbf{Post-Processing and Tape Validation.} To optimize TTS quality at inference, we merge and resegment continuous agent text into clause-level granularity using a punctuation model \cite{guhr-EtAl:2021:fullstop}, which is the finest granularity that preserves prosodic quality. This resegmentation is applied independently to each contiguous span of agent text with only state transition tokens between them, with \verb|[C.SPEAK]| tokens inserted at clause boundaries. We then validate every resulting sequence against the FSM's transition constraints (e.g., the \verb|SPEAK| state emits only \verb|[S.LISTEN.*]| or \verb|[C.SPEAK]|) to ensure the completeness of the entire pipeline.

\subsection{Human-Agent Data Transformation}
To preserve the LLM's intrinsic semantic abilities, we leverage broad-domain HA text dialogues. However, as textual HA datasets are primarily written-style and thus unsuitable for a spoken dialogue agent, we convert them into the FSM tape format through a three-stage pipeline: (1) rule-based filtering, (2) LLM-driven stylistic rewriting, and (3) final filtering and tape serialization.

We first apply rule-based cleaning to the raw text dialogues, which strips HTML markup, validates role alternation, and removes turns dominated by code, mathematical formulas, or excessively long contexts. We then prompt an LLM with a two-phase prompt (Appendix \ref{sec:appendix_prompts}) that first judges suitability for spoken delivery, then rewrites the surviving dialogues into a natural spoken style. A final filtering pass discards turns exceeding spoken-utterance length thresholds or containing residual Markdown artifacts, yielding a distilled set of high-quality dialogues.

Finally, we segment user and agent utterances into clauses with the same punctuation model as in HH transformation, then insert \verb|[S.SPEAK]| and \verb|[S.LISTEN.N]| between alternating utterances and \verb|[C.LISTEN]|/\verb|[C.SPEAK]| between clauses within the same utterance. This establishes a trivial yet schema-compliant turn-taking structure, allowing the HA data to focus its supervisory signal on semantic content.

\section{Model}

\subsection{FSM Modules}
In the absence of a publicly available NFSM implementation, we build the entire system from scratch, integrating intrinsically multilingual perception, cognitive, and motor modules optimized for real-time interaction. To examine the effect of perception granularity, we evaluate two streaming ASR alternatives, both built on Whisper-Turbo \cite{radford2022robustspeechrecognitionlargescale} to isolate granularity from backbone: a word-level streaming ASR, SimulStreaming \cite{machacek-polak-2025-simultaneous}, and an IPU-level streaming ASR coupling Silero VAD \cite{Silero_VAD} with Faster-Whisper.\footnote{https://github.com/SYSTRAN/faster-whisper} This contrast allows us to explore the fundamental trade-off between the precise temporal resolution afforded by fine-grained streaming and the semantic coherence required for aggressively fragmented inputs. For the motor module, we adopt Kokoro TTS,\footnote{https://github.com/hexgrad/kokoro} whose Time-to-First-Audio on clause-level inputs outperforms many streaming alternatives without sacrificing prosodic quality. Other proposed techniques for FSM are available in Appendix \ref{sec:appendix_fsm_design}.

\subsection{Source-Aware Calibrated Loss}
Optimizing the FSM model presents two challenges: mitigating the severe class imbalance inherent in turn-taking behaviors and jointly optimizing two distinct capabilities. To this end, we propose the SAC Loss, a dual-purpose objective designed to calibrate the long-tail token distribution and decouple learning targets across diverse data sources.

During fine-tuning, we adopt a selective masking strategy. Specifically, we mask the prompts and user text, restricting the loss computation exclusively to the agent response tokens and state transition tokens. Let $I = \{1, \dots, N\}$ denote the target token indices for a training instance. We partition $I$ along two orthogonal dimensions: by token type into response tokens ($R$) and state transition tokens ($T$); and by data source into Human-Human ($HH$) and Human-Agent ($HA$) tokens.

To achieve distributional calibration, we tackle the extreme class imbalance among state transition tokens. Continuation tokens representing ongoing listening or speaking appear exponentially more frequently than discrete state-switching tokens (see Appendix~\ref{sec:appendix_token_stats} for the distribution). Employing a standard cross-entropy objective naturally marginalizes these low-probability yet operationally critical transitions. Therefore, we incorporate Logit Adjustment \cite{menon2021longtail} specifically for the state transition tokens to rectify this skewed prior, while retaining the standard cross-entropy loss for the response tokens. The per-component losses for an input $x$ and target label $y$ are defined as follows:
$$
\ell^{\mathrm{CE}}(x, y) = -\log \frac{\exp(f_{y}(x))}{\sum_{v} \exp(f_v(x))}
$$
$$
\ell^{\mathrm{LA}}(x, y) = -\log \frac{\exp\bigl(f_{y}(x) + \tau \log \pi_{y}\bigr)}{\sum_{v} \exp\bigl(f_v(x) + \tau \log \pi_v\bigr)}
$$
where $f(\cdot)$ denotes the model logits, $\pi_y$ is the prior probability of class $y$, and $\tau$ is the adjustment temperature. The prior $\pi$ is computed exclusively over the restricted state-transition vocabulary to avoid overweighting. Accordingly, the loss $\ell_i$ for the $i$-th token is formulated as:
$$
\ell_i =
\begin{cases}
    \ell^{\mathrm{CE}}(x_i, y_i), & \text{if } i \in R \\
    \ell^{\mathrm{LA}}(x_i, y_i), & \text{if } i \in T
\end{cases}
$$

To achieve source-aware specialization, we introduce a weighting mechanism that dynamically adjusts the importance of each token based on its type and origin. The overall training loss $\mathcal{L}$ is computed as a weighted average over all unmasked tokens:
$$
\mathcal{L} = \frac{\sum_{i \in I} w_i \ell_i}{\sum_{i \in I} w_i}
$$
where the weight $w_i$ assigned to the $i$-th token is defined as:
$$
w_i =
\begin{cases}
    \alpha, & \text{if } i \in (HA \cap R) \cup (HH \cap T) \\
    1 - \alpha, & \text{if } i \in (HA \cap T) \cup (HH \cap R)
\end{cases}
$$
in which $\alpha \in (0.5, 1]$ is a hyperparameter controlling the degree of source emphasis. By assigning a higher weight $\alpha$ to the subsets $(HH \cap T)$ and $(HA \cap R)$, the SAC Loss explicitly decouples the optimization trajectories: it forces the model to learn realistic turn-taking dynamics predominantly from human-human interactions, while refining its semantic capabilities primarily through human-agent data.

\section{Experiments}
\subsection{Settings}
\label{sec:settings}
For HH spoken dialogues, we adopt the Switchboard \cite{10.5555/1895550.1895693} and Fisher \cite{cieri-etal-2004-fisher} corpora. For the HA text dialogues, we adopt the ShareGPT dataset\footnote{https://huggingface.co/datasets/RyokoAI/ShareGPT52K} and utilize Qwen3-32B \citep{yang2025qwen3technicalreport} for stylistic rewriting.
 As the foundation model, we adopt the open-source lightweight Qwen3-4B (model scale analysis in Appendix~\ref{sec:appendix_hyper_scaling}) and extend its tokenizer with state transition tokens, the silence token \verb|<SIL>|, and the role-prefix token \verb|<user>|. For the data mixture, we set HH:HA token-volume ratio as $1{:}1$, with Fisher:Switchboard mixed at $4{:}1$ within HH. For the SAC loss, we set $\tau=1$ and $\alpha=0.6$. Hyperparameter analyses are provided in Appendix~\ref{sec:appendix_hyper_scaling}, while full training details, dataset splits, and First Token Emission Delay (FTED) latency are in Appendix \ref{sec:appendix_training} and \ref{sec:appendix_latency}.

Considering the original training dataset of NFSM is not publicly available, we reproduce their training data to establish a baseline representative of the conventional synthetic data approach. Specifically, GPT-4-Turbo \cite{openai2024gpt4technicalreport} generates 1,500 synthetic dialogues using the published NFSM prompts, which we serialize via the same procedure as our HA transformation, with one exception---incomplete-utterance markers (\verb|<NOT_FINISHED>|, trailing ellipses) are mapped to \verb|[S.LISTEN.I]|. We then fine-tune a separate Qwen3-4B model on this synthetic tape, which serves as our primary baseline. More details are available in Appendix \ref{sec:appendix_baseline}.

Our evaluation focuses on turn-taking proficiency and semantic capability. Turn-taking is measured on HH corpora as the ground truth for natural dynamics: at each teacher-forced prediction step, we treat the output as an independent binary classification for each state transition token type, compute per-type F1 within each evaluation set, average across types within each set, then average across Switchboard and Fisher to obtain the reported F1. For semantic evaluation, we adopt VoiceBench \cite{chen2024voicebenchbenchmarkingllmbasedvoice} following \citet{xiang-etal-2025-understanding}, excluding IFEval to better match a purely acoustic system and averaging over SD-QA, MMSU, OpenBookQA, and AdvBench subsets as a result. All test inputs are prepended with the same FSM prompt. For fine-tuned models, evaluation audio is transcribed by their respective perception modules and interleaved with state transition tokens. For the zero-shot foundation model upper bound, audio is transcribed by Whisper-Turbo into continuous plain text, preventing degradation from context fragmentation.

\begin{table}[t]
\centering
\small
\setlength{\tabcolsep}{5pt}
\begin{tabular}{@{}lcccc@{}}
\toprule
 & \multicolumn{2}{c}{\textbf{Faster-Whisper}} & \multicolumn{2}{c}{\textbf{SimulStreaming}} \\
 & \multicolumn{2}{c}{\textit{(IPU-level)}} & \multicolumn{2}{c}{\textit{(word-level)}} \\
\cmidrule(lr){2-3} \cmidrule(lr){4-5}
\textbf{Model} & \textbf{F1} $\uparrow$ & \textbf{VB} $\uparrow$& \textbf{F1} $\uparrow$& \textbf{VB} $\uparrow$\\ \midrule
NFSM           & 0.3436 & 53.57 & 0.3025 & 44.24 \\
Ours  & \textbf{0.6498} & \textbf{62.14} & \textbf{0.6404} & \textbf{54.80} \\ \midrule
\textit{$\Delta$ vs. NFSM} & \textit{+0.3062} & \textit{+8.57} & \textit{+0.3379} & \textit{+10.56} \\
\bottomrule
\end{tabular}
\caption{Overall performance across method and perception module, where VB denotes VoiceBench.}
\label{tab:overall}
\end{table}

\begin{figure*}[t]
  \includegraphics[width=0.45\linewidth]{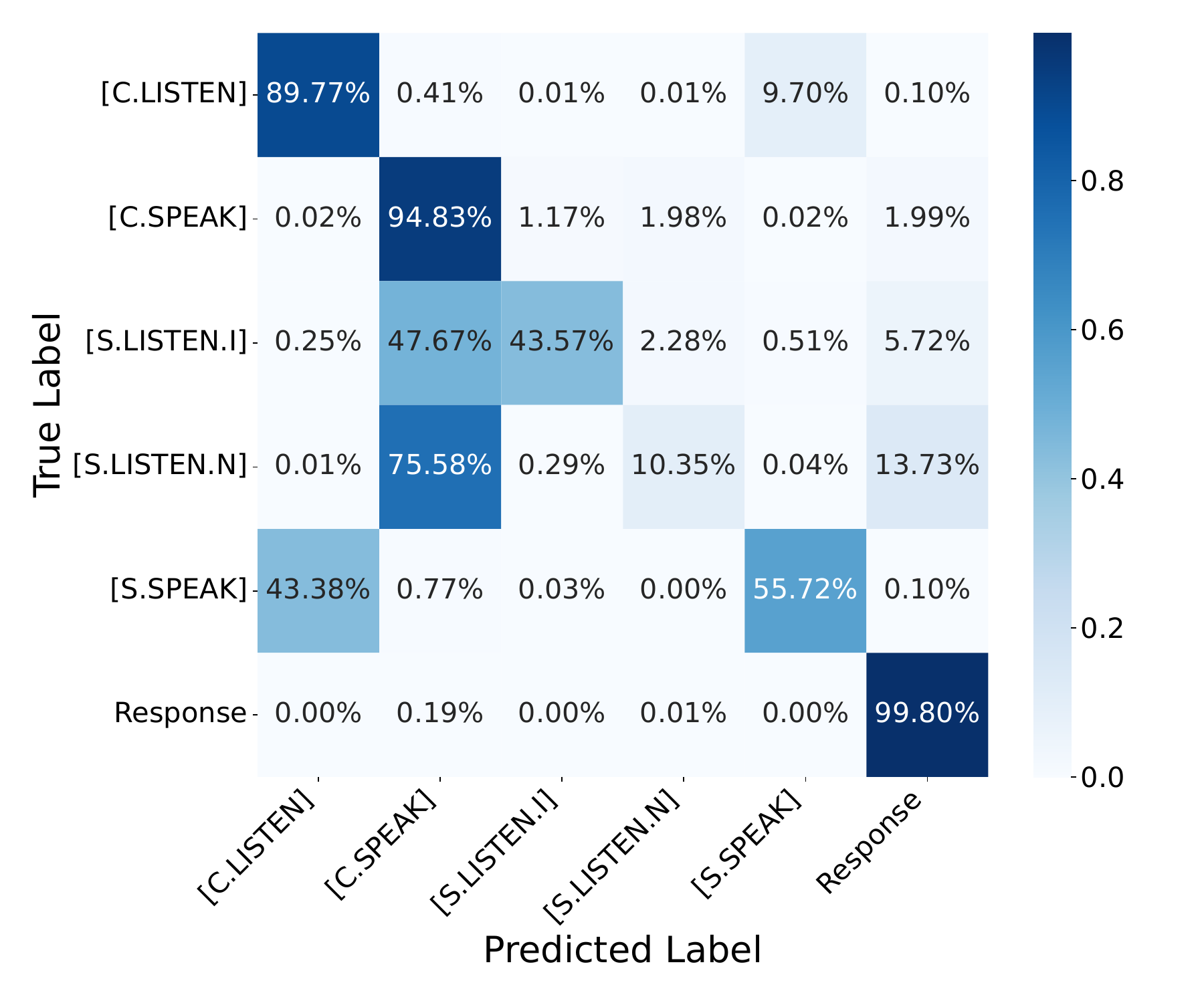} \hfill
  \includegraphics[width=0.45\linewidth]{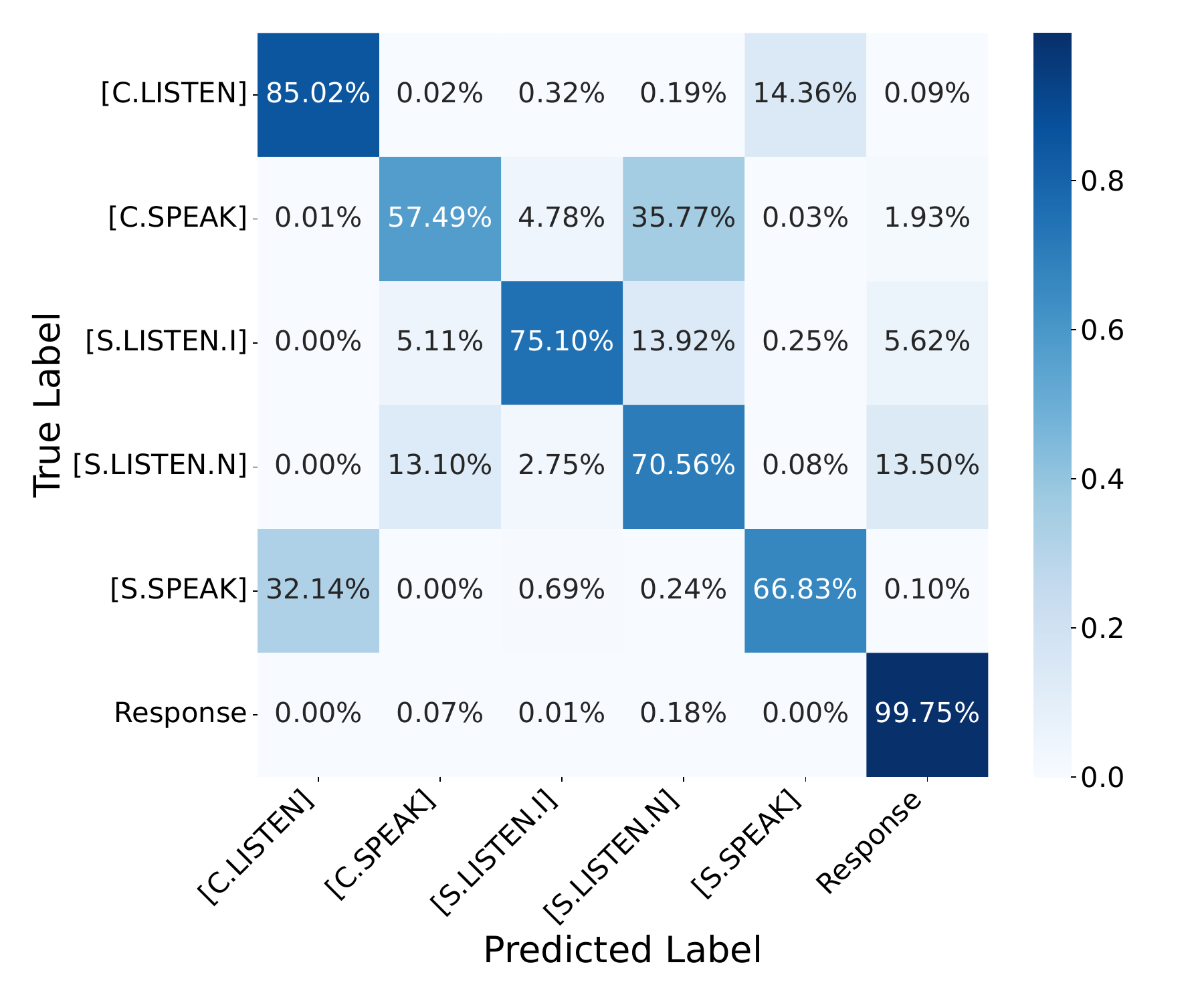}
  \caption {Confusion matrices for state transition and response tokens on Switchboard with the Faster-Whisper-based model: standard cross-entropy (left) versus SAC loss (right).}
  \label{fig:confusion_matrix}
\end{figure*}

\subsection{Main Results}

To ensure fair comparison, we control the total training token volume for our models to match the NFSM baseline. As shown in Table \ref{tab:overall}, our model significantly outperforms NFSM under both perception configurations. Comparing the two variants of our model, the Faster-Whisper variant achieves turn-taking performance comparable to the SimulStreaming variant while attaining substantially higher semantic capability. This gap stems from the hyper-granular nature of SimulStreaming outputs: user textual content is heavily fragmented by frequently interleaved state transition tokens, challenging the LLM's semantic comprehension (details in Appendix \ref{sec:appendix_token_stats}). These findings reveal a critical architectural trade-off in perception module design: finer granularity offers more frequent opportunities to manage interruptions but risks degrading semantic prowess.

\begin{table}[t]
\centering
\small
\setlength{\tabcolsep}{8pt}
\begin{tabular}{@{}lrr@{}}
\toprule
\textbf{Configuration} & \textbf{F1} $\uparrow$ & \textbf{VB} $\uparrow$ \\ \midrule
Ours & 0.6498 & 62.14 \\ \midrule
\multicolumn{3}{@{}l}{\textit{Ablation on Loss Function}} \\
\quad w/o SAC loss & 0.6105 & 62.19 \\ \midrule
\multicolumn{3}{@{}l}{\textit{Ablation on Training Data (w/o SAC loss)}} \\
\quad HH only        & 0.6182 & 18.30 \\
\quad HA only        & 0.4170 & 63.15 \\
\quad Synthetic only (NFSM) & 0.3436 & 53.57 \\ \bottomrule
\end{tabular}
\caption{Ablation study on the SAC loss and training data composition under Faster-Whisper perception.}
\vspace{-3mm}
\label{tab:ablation_faster}
\end{table}

\begin{table}[t]
\centering
\small
\setlength{\tabcolsep}{8pt}
\begin{tabular}{@{}lrr@{}}
\toprule
\textbf{Configuration} & \textbf{F1} $\uparrow$ & \textbf{VB} $\uparrow$ \\ \midrule
\multicolumn{3}{@{}l}{\textit{Semantic Upper Bound}} \\
\quad Qwen3-4B (zero-shot) & 0.1200 & 64.93 \\ \midrule
\multicolumn{3}{@{}l}{\textit{End-to-End Reference}} \\
\quad Moshi & --- & 27.36 \\ \midrule
\multicolumn{3}{@{}l}{\textit{Our FSM Model}} \\
\quad Token-matched subset & 0.6498 & 62.14 \\
\quad Proportionate upsampling & 0.6539 & 64.84 \\ \bottomrule
\end{tabular}
\caption{Performance of our FSM model scaled to the proportionate upsampling mixture, alongside the zero-shot semantic upper bound and end-to-end reference.}
\label{tab:full_data_faster}
\vspace{-3mm}
\end{table}

\begin{table*}[t]
\centering
\small
\footnotesize
\setlength{\tabcolsep}{3pt}
\begin{tabular}{@{}lcccccccccccc@{}}
\toprule
 & \multicolumn{8}{c}{\textbf{Full-Duplex-Bench v1.0}} & \multicolumn{4}{c}{\textbf{Full-Duplex-Bench v1.5}} \\
\cmidrule(lr){2-9}\cmidrule(lr){10-13}
 & \multicolumn{2}{c}{Pause Handling} & \multicolumn{3}{c}{Backchannel} & Smooth Turn & \multicolumn{2}{c}{User Interruption} & \multicolumn{4}{c}{Overlap Handling} \\
\cmidrule(lr){2-3}\cmidrule(lr){4-6}\cmidrule(lr){7-7}\cmidrule(lr){8-9}\cmidrule(lr){10-13}
\textbf{Data} & Synthetic & Candor & \multicolumn{3}{c}{ICC} & Candor & \multicolumn{2}{c}{Synthetic} & U-INTR & U-BC & T-OTH & BKG \\
\textbf{Model} & TOR$\downarrow$ & TOR$\downarrow$ & TOR$\downarrow$ & Freq$\uparrow$ & JSD$\downarrow$ & TOR$\uparrow$ & TOR$\uparrow$ & GPT-4o$\uparrow$ & RESP$\uparrow$ & RSM$\uparrow$ & RSM$\uparrow$ & RSM$\uparrow$ \\
\midrule
Moshi        & 0.985 & 0.980 & 1.000 & 0.001 & 0.957 & \textbf{0.941} & \textbf{1.000} & 0.765 & 0.50 & 0.06 & 0.19 & 0.07 \\
Freeze-Omni  & \underline{0.642} & \underline{0.481} & 0.636 & 0.001 & 0.997 & 0.336 & 0.867 & \textbf{3.615} & \textbf{0.72} & 0.80 & 0.25 & 0.25 \\
Gemini Live       & \textbf{0.255} & \textbf{0.310} & \textbf{0.091} & \underline{0.012} & \underline{0.896} & 0.655 & \underline{0.891} & 3.376 & 0.33 & \underline{0.93} & \textbf{0.99} & \underline{0.30} \\
Ours   & 0.825 & 0.653 & \underline{0.127} & \textbf{0.119} & \textbf{0.764} & \underline{0.891} & 0.875 & \underline{3.589} & \underline{0.69} & \textbf{0.98} & \underline{0.44} & \textbf{0.36} \\
\bottomrule
\end{tabular}
\caption{Full-Duplex-Bench results for our proportionate upsampling model, restricted to the systems common to both benchmarks. U-INTR: user interruption; U-BC: user backchannel; T-OTH: talking to others; BKG: background speech; RESP: respond rate; RSM: resume rate.}
\label{tab:fdb_merged}
\end{table*}

\subsection{Ablation Study}

For the ablation study, we first examine the effectiveness of the SAC loss. As shown in Table \ref{tab:ablation_faster}, integrating the SAC loss consistently improves turn-taking proficiency while preserving semantic capability at a comparable level. Its central advantage, however, lies in mitigating the severe class imbalance among state transition tokens. Figure \ref{fig:confusion_matrix} presents the confusion matrices for state transition and response tokens on Switchboard with the Faster-Whisper-based model. Specifically, each decoding step is formulated as an independent classification task and percentages are normalized against the ground-truth labels. The results clearly show that the SAC loss yields notable gains in the prediction accuracy of state-switch tokens, which are operationally critical yet rare under the standard cross-entropy objective.

Furthermore, we evaluate the impact of data mixture strategies. Under identical loss configurations, the model trained solely on synthetic data is substantially outperformed by the model trained on the combined HH and HA corpora. For turn-taking, this gap arises from two factors: current synthetic generation strategies fail to adequately simulate the fine-grained acoustic dynamics of real human dialogues, and synthetic transcripts cannot replicate the specific temporal granularities and recognition artifacts of the perception module. For semantic preservation, authentic HA dialogues prove more effective than synthetic transcripts that attempt to distill both capabilities simultaneously. Finally, while training exclusively on either HH or HA enhances one capability in isolation, neither corpus alone improves both. We observe highly consistent trends when substituting Faster-Whisper with SimulStreaming as the perception module, with detailed results provided in Appendix \ref{sec:appendix_simul_ablation}.

Beyond the turn-taking F1, we further assess interruption handling under both machine-interrupts-user and user-interrupts-machine settings, following the protocol of NFSM. Our model substantially outperforms NFSM in both directions (Appendix~\ref{sec:appendix_interruption}).

\subsection{Scaling to the Proportionate Upsampling Mixture}
While the preceding experiments operated under a strict token-volume constraint (referred to as the token-matched subset), we now investigate model performance when scaling training to its maximum extent. To strictly preserve the established blending ratio across the constituent corpora, we construct a proportionally upsampled mixture by repeatedly oversampling the datasets to match the volume of the largest split (detailed in Appendix \ref{sec:appendix_upsampling}). As shown in Table \ref{tab:full_data_faster}, scaling yields further improvements in both turn-taking proficiency and semantic capability. To rigorously assess the extent of semantic recovery, we introduce a strong upper bound: the original Qwen3-4B in zero-shot setting. Remarkably, despite being trained to handle complex full-duplex turn-taking on a fragmented FSM tape, our Faster-Whisper-based model recovers the intrinsic semantic capability of the underlying LLM, closing the gap to within 0.09 points on VoiceBench. As an additional reference, the end-to-end Moshi attains 27.36 on the same VoiceBench subset as reported by \citet{chen2024voicebenchbenchmarkingllmbasedvoice}, which is lower than our FSM model and consistent with prior observations that end-to-end full-duplex architectures often trade off semantic capability.

We further evaluate the proportionate upsampling model on Full-Duplex-Bench \cite{lin2025fdb_v1, lin2025fdb_v15}, an architecture-agnostic benchmark independent of both our token scheme and training distribution (Table~\ref{tab:fdb_merged}; complete results in Appendix~\ref{sec:appendix_fdb}). The most substantial gains are observed in backchanneling: our model achieves the best human-aligned timing and highest frequency with a low takeover rate on v1.0, while attaining the highest resume rate on v1.5 user backchannel. We attribute this to the event-guided tape and the SAC Loss. The former supervises what kind of overlap occurred rather than merely whether speech is present, while the latter prevents the rare switch-type tokens from being washed out by dominant response tokens. The model remains highly competitive across most remaining dimensions, ranking first or second in smooth turn-taking, user-interruption semantic quality, and all four v1.5 overlap scenarios. A notable exception is pause handling, where our takeover rates trail Freeze-Omni and Gemini Live, indicating a more aggressive strategy in reclaiming the floor. However, this behavior reflects an inherent turn-taking trade-off rather than a strict deficiency. For instance, the conservative Freeze-Omni excels at pause handling but compromises smooth turn-taking. Overall, our approach outperforms every baseline across the majority of the twelve metrics (Moshi $10/12$, Freeze-Omni $8/12$, Gemini Live $7/12$).

\section{Conclusion}
In this work, we presented a decoupled data approach for FSM-based full-duplex dialogue, in which turn-taking is acquired from real HH spoken dialogues while semantic behavior is shaped through configurable HA text dialogues. We operationalized this approach with a fully rule-based event-guided transformation that serializes HH spoken dialogues into causal FSM tapes without LLM-generated annotations, and introduced the SAC Loss to jointly calibrate the long-tailed distribution of state transition tokens and channel each data source toward the capability it best supervises. Experiments show that our approach substantially improves turn-taking proficiency over the synthetic-data baseline while recovering the foundation LLM's semantic capability to the zero-shot upper bound, demonstrating that decoupling data sources by capability provides a scalable recipe for FSM-based full-duplex dialogue.

\section*{Limitations}
Our study has four primary limitations.
First, serializing two parallel audio channels into a single causal tape inevitably compresses fine-grained continuous-time information into a discrete token order, leading to information loss despite our enforcement of temporal causality.
Second, paralinguistic information---prosody, emotion, and voice quality---is not representable on the text-mediated tape, and the expressiveness of the rendered speech is bounded by the off-the-shelf TTS module, which is inherent to the original NFSM formulation. In this work, we focus on turn-taking naturalness and leave paralinguistic cues and a perceptual study to future.
Third, our HH corpora retain the agent channel's original GT utterance content as spoken by ordinary human participants, which does not necessarily align with the response style expected of a voice assistant. Refining the semantic content of the agent channel could yield a higher-quality training signal, but risks disturbing the temporal alignment between channels and thus the turn-taking dynamics we aim to preserve. We therefore leave the agent channel unedited and view its controlled refinement as a promising direction for future work.
Fourth, our corpora cover only two-party dialogue and a question-answering assistant role. Extending the HH data to multi-party settings and the HA data to role-conditioned dialogues \cite{mitra2026adaptiveturntakingrealtimemultiparty} is left to future work.

\section*{Acknowledgements}
This work was supported by JST SPRING (Grant Number JPMJSP2110) and JSPS (Grant Number JP23K28144). We thank Professor Tatsuya Kawahara for his assistance in obtaining access to the Fisher corpus.


\bibliography{custom}

\appendix

\section{Data Construction Details}
\label{sec:appendix_data}

\begin{table*}[t]
\centering
\small
\begin{tabular}{@{}lll@{}}
\toprule
\textbf{Parameter} & \textbf{Specified in NFSM} & \textbf{Our reproduction} \\
\midrule
Generation LLM & Yes (\texttt{gpt-4-turbo-2024-04-09}) & Identical\\
Decoding settings & No & Default sampling \\
\# dialogues & Yes (1,500 series) & Identical\\
Prompt template & Yes (their Appendix A) & Identical \\
\texttt{\{num\_rounds\}} & No (placeholder only) & Uniform over 9--11 \\
Scenario-to-round assignment & No (placeholder only) & Random round with no collision \\
\texttt{\{response\_word\_count\}} & No (placeholder only) & Uniform over 100--150 \\
\texttt{\{interrupted\_response\_word\_count\}} & No (placeholder only) & Uniform over 20--50 \\
Topic pool & Partial (``hundreds'', ``random'') & \makecell[l]{110 topics generated with \texttt{gpt-4-turbo}\\\texttt{-2024-04-09}} \\
Dialogue serialization & No & As described in Section~\ref{sec:settings}\\
\bottomrule
\end{tabular}
\caption{Reconstruction parameters of the NFSM synthetic-data baseline. For each parameter we indicate whether it is specified in the original NFSM paper and the setting adopted in our reproduction.}
\label{tab:baseline_repro}
\end{table*}

\subsection{NFSM Baseline Reproduction Details}
\label{sec:appendix_baseline}

Because the original NFSM training corpus is not publicly available, our synthetic-data baseline is reconstructed from the specifications given in the NFSM paper. For full transparency, Table~\ref{tab:baseline_repro} documents every parameter of this reconstruction, indicating for each whether it is specified in the original paper and, where it is not, the setting we adopt. We reproduce the original work as faithfully as possible, keeping its most critical components identical to the original, including the generation LLM and the full prompt template. A few minor parameters such as the unfilled placeholders in the published prompt are not disclosed, so an exact match is impossible and we adopt reasonable settings.

\subsection{State Transition Token Statistics}
\label{sec:appendix_token_stats}

Table \ref{tab:token_stats} reports the token type distribution of the Switchboard training tapes under the two perception modules, placed side by side with that of the reconstructed NFSM synthetic tapes for reference. The distributions exhibit three notable patterns.

First, state transition tokens follow a markedly long-tailed distribution. Within the restricted state transition vocabulary, continuation tokens jointly account for $72.78\%$ and $80.45\%$ of all state transition tokens under the Faster-Whisper and SimulStreaming perception modules respectively, whereas the operationally critical state-switch tokens occupy a significantly smaller proportion. This skew motivates our use of logit adjustment over the state transition tokens in the SAC loss.

Second, the choice of perception module reshapes the distribution in a predictable direction. Switching from the IPU-level Faster-Whisper to the word-level SimulStreaming raises the share of \verb|[C.LISTEN]| from $8.75\%$ to $13.77\%$. The increased interleaving of \verb|[C.LISTEN]| within user content quantitatively reflects the fragmentation phenomenon and offers a partial explanation for the larger semantic gap observed under SimulStreaming.

Third, the synthetic tapes exhibit a markedly more extreme imbalance than the real HH data: response tokens account for $88.38\%$ of the tape, leaving only $11.62\%$ for all state-transition tokens, of which the operationally critical state-switch tokens (\verb|[S.LISTEN.I]|, \verb|[S.LISTEN.N]|, and \verb|[S.SPEAK]|) constitute merely $2.40\%$. In contrast, the real HH tapes allocate a substantially larger share to state-switch tokens of $7.26\%$ under Faster-Whisper and $5.88\%$ under SimulStreaming. This gap reflects that LLM-generated transcripts, being predominantly structured and non-overlapping, seldom reproduce the frequent floor changes, backchannels, and interruptions of natural conversation.

\begin{table}[t]
\centering
\small
\begin{tabular}{@{}lrrr@{}}
\toprule
 & \textbf{} & \multicolumn{2}{c}{\textbf{Real HH}} \\ \cmidrule(l){3-4} 
\textbf{Token} & \textbf{Synthetic} 
 & \textbf{\begin{tabular}[b]{@{}r@{}}Faster-\\Whisper\end{tabular}}
 & \textbf{\begin{tabular}[b]{@{}r@{}}Simul-\\Streaming\end{tabular}} \\ \midrule
\verb|[C.LISTEN]|   &  0.91\% &  8.75\% & 13.77\% \\
\verb|[C.SPEAK]|    &  8.31\% & 10.63\% & 10.42\% \\
\verb|[S.LISTEN.I]| &  0.17\% &  0.77\% &  0.53\% \\
\verb|[S.LISTEN.N]| &  0.99\% &  2.83\% &  2.38\% \\
\verb|[S.SPEAK]|    &  1.24\% &  3.66\% &  2.97\% \\
Response            & 88.38\% & 73.37\% & 69.93\% \\ \midrule
Total               & 100.00\% & 100.00\% & 100.00\% \\ \bottomrule
\end{tabular}
\caption{Token-type ratio across the NFSM synthetic tapes and the Switchboard training tapes under two perception modules.}
\label{tab:token_stats}
\end{table}

\begin{table*}[t]
\centering
\small
\begin{tabular}{@{} l rrrr rrr rr @{}}
\toprule
\multirow{2}{*}{\textbf{Configuration}} & \multicolumn{4}{c}{\textbf{Train (k)}} & \multicolumn{3}{c}{\textbf{Validation (k)}} & \multicolumn{2}{c}{\textbf{Test (k)}} \\
\cmidrule(lr){2-5} \cmidrule(lr){6-8} \cmidrule(l){9-10}
& \textbf{FSH} & \textbf{SWBD} & \textbf{SGPT} & \textbf{SYN} & \textbf{FSH} & \textbf{SWBD} & \textbf{SYN} & \textbf{FSH} & \textbf{SWBD} \\
\midrule
Baseline & -- & -- & -- & \num{1344} & -- & -- & \num{165} & \num{5600} & \num{444} \\
Ours (Token-matched) & \num{538} & \num{135} & \num{672} & -- & \num{691} & \num{236} & -- & \num{5600} & \num{444} \\
Ours (Scaled Mixture) & \num{21792} & \num{5448}$^*$ & \num{27240}$^\dagger$ & -- & \num{691} & \num{236} & -- & \num{5600} & \num{444} \\
\bottomrule
\end{tabular}
\caption{Dataset statistics across different configurations, measured in thousands (k) of tokens. Abbreviations used are FSH (Fisher), SWBD (Switchboard), SGPT (ShareGPT), and SYN (Synthetic). For the scaled mixture, we apply proportional oversampling to maintain the established data distribution. The superscript symbols indicate datasets that are repeatedly sampled from their natural volume: ($^*$) base size $\num{1551}\text{k} \times 3.51$, and ($^\dagger$) base size $\num{2966}\text{k} \times 9.18$.}
\label{tab:data_statistics}
\end{table*}

\subsection{Merging Rules at Segment Boundaries}
\label{sec:appendix_boundary}

Concatenating two adjacent segments during serialization may produce two consecutive state transition tokens at their boundary, which must be collapsed into a single token to align with the rule of FSM.

Rather than enumerating $14 \times 14$ event pairs or $5 \times 5$ token pairs, we observe that the merged token is fully determined by the FSM state immediately before the first token fires and immediately after the second token fires, reducing the analysis to four state-pair cases:

\begin{itemize}
    \item (\verb|LISTEN|, \verb|LISTEN|) $\rightarrow$ \verb|[C.LISTEN]|: the FSM remains in \verb|LISTEN|.
    \item (\verb|SPEAK|, \verb|SPEAK|) $\rightarrow$ \verb|[C.SPEAK]|: the FSM remains in \verb|SPEAK|.
    \item (\verb|LISTEN|, \verb|SPEAK|) $\rightarrow$ \verb|[S.SPEAK]|: a switch from listening to speaking.
    \item (\verb|SPEAK|, \verb|LISTEN|) $\rightarrow$ \verb|[S.LISTEN.*]|: a switch from speaking to listening, where the N/I subtype is inherited from whichever token in the pair is \verb|[S.LISTEN.*]|, with \verb|[S.LISTEN.I]| taking priority when both are present.
    \end{itemize}

The (\verb|SPEAK|, \verb|LISTEN|) case warrants further justification, as it is the only case requiring subtype inheritance. The first token in this case must be \verb|[C.SPEAK]| or \verb|[S.LISTEN.*]|, and the second must be \verb|[C.LISTEN]| or \verb|[S.LISTEN.N]|. Among these, the pair \verb|[C.SPEAK]| + \verb|[C.LISTEN]| appears superficially well-formed but never arises in valid serializations, so the merge can always recover an N/I subtype from one side of the pair. To see why, note that a segment ending in \verb|[C.SPEAK]| can only originate from BC or BI initiated by the user, or FTI initiated by the agent (Table~\ref{tab:Serialization_rule}); in all three cases the floor holder at the segment's end is the agent. Conversely, a segment beginning with \verb|[C.LISTEN]| can only originate from C, P, or G initiated by the user, all of which require the user to be the floor holder at the segment's start (recall that the initiator of a G segment is the preceding floor holder). These two requirements are mutually inconsistent, ruling out the pair. 

Besides, as a related adjustment, when a T immediately follows a G, its leading \verb|[S.LISTEN.N]| is demoted to \verb|[C.LISTEN]|, as the preceding G has already transitioned the FSM into \verb|LISTEN|.

\subsection{Proportionate Upsampling Mixture}
\label{sec:appendix_upsampling}

For the scaling experiments, we construct a proportionally upsampled mixture to maximize the training signal while strictly maintaining the established data distribution. Since the constituent corpora possess disparate natural volumes, datasets with smaller token counts are oversampled to adhere to the predefined ratios. Detailed statistics for each configuration, including base volumes, oversampling multipliers, and final token counts, are provided in Table \ref{tab:data_statistics}. This approach ensures that the model benefits from the full extent of available data without biasing the optimization toward any single data source.

\subsection{Prompts}
\label{sec:appendix_prompts}
\lstdefinestyle{promptstyle}{
  backgroundcolor=\color{black!5},   
  basicstyle=\ttfamily\small,      
  breaklines=true,                 
  rulecolor=\color{black},         
  captionpos=b,                    
  keywordstyle=\color{blue},       
  morekeywords={Human, AI},         
  xleftmargin=0pt,
}

The prompt for filtering and stylistic rewriting HA data is shown as follows:
\begin{tcolorbox}[
  colback=gray!5,  
  colframe=gray!50,
  title=Prompt,
  fonttitle=\bfseries,
  breakable,       
  boxrule=0.5pt   
]
\small\ttfamily 
You are an expert linguistic editor specializing in dialogue style transfer. Your task is twofold: first, EVALUATE whether the input text is suitable for conversion into a casual spoken dialogue; second, ONLY IF suitable, EXECUTE the conversion.\\

\#\#\# PHASE 1: EVALUATION (Strict Filtering)\\
Before converting, analyze the input content. You must return an empty list `[]` for the 'conversations' field if the input is not suitable for spoken dialogue, for example but not constrain to the following 'UNSUITABLE' categories:\\
1. **Code \& Technical Data**: Contains programming code, stack traces, logs, JSON/XML, SQL queries, or shell commands.\\
2. **Strict structured text**: Contains mathematical formulas (LaTeX), chemical equations, or rigid data tables.\\
3. **Document text**: Contains document input which is impossible to happen in dialogue.\\
4. **Non-Dialogue / Fragments**: The input is just a list of keywords, a disjointed sentence fragment without context, or incoherent noise.\\
5. **Competitor Identity / Specific AI Models**: The text mentions 'OpenAI', 'ChatGPT', or the model refers to itself as such.\\

\#\#\# PHASE 2: CONVERSION (Only if Phase 1 passes)\\
If the content is suitable, convert it into a natural spoken-style conversation.
Guidelines for Spoken Style:\\
- Strictly REMOVE all Markdown syntax (e.g., *, \#, -). Do not use bullet points or numbered lists; use connecting words like 'First', 'Next', 'Finally' instead.\\
- Simplify complex sentence structures and content to be more conversational.\\
- Maintain the original intent and core information.\\

\#\#\# CONSTRAINTS:\\
1. Keep the original language of each utterance unchanged.\\
2. Keep the original 'id' strictly unchanged.\\
3. Ensure the speaker roles and the number of turns remain exactly the same.\\
4. **CRITICAL**: If the content matches the UNSUITABLE criteria, the output must be `[]`.\\
5. Ensure the output follows the strict JSON schema provided.\\
\end{tcolorbox}

The prompt of FSM is shown as follows:
\begin{tcolorbox}[
  colback=gray!5,  
  colframe=gray!50,
  title=Prompt,
  fonttitle=\bfseries,
  breakable,       
  boxrule=0.5pt   
]
\small\ttfamily 
\textless{}|im\_start|\textgreater{}user\\
You are a helpful voice assistant.\\
You will work as a Finite State Machine to handle state transitions between LISTEN and SPEAK states.\\
State-transition tokens: "[S.SPEAK]" means "switch to SPEAK"; "[S.LISTEN.INTERRUPT]" means "switch to LISTEN because of being interrupted"; "[S.LISTEN.NATURAL]" means "switch to LISTEN because of natural end"; "[C.SPEAK]" means "continue to SPEAK"; "[C.LISTEN]" means "continue to LISTEN".\\
For content between State-transition tokens, the user's content are prefixed with "\textless{}user\textgreater{}" and your own content has no prefix. "\textless{}SIL\textgreater{}" represents a silence of \{\{SILENCE\_TOKEN\_DUR\}\} seconds.\\
Please generate a tape fragment of the Finite State Machine.\textless{}|im\_end|\textgreater{}\\
\textless{}|im\_start|\textgreater{}assistant\\
\textless{}think\textgreater{}\\

\textless{}/think\textgreater{}
\end{tcolorbox}

\section{Additional Experimental Settings}
\label{sec:appendix_settings}

This section supplements the general settings in the main text with details on training hyperparameters, dataset partitioning, and the temporal thresholds used in the FSM and our event-guided transformation pipeline.

\subsection{Training Hyperparameters and Dataset Partitioning}
\label{sec:appendix_training}

Fine-tuning is conducted on four NVIDIA RTX A6000 GPUs with batch size $256$, optimized with AdamW (weight decay $0.01$) under a linear schedule with warmup, peak learning rate $1.0\times10^{-5}$, and warmup ratio $0.03$. Each instance has a context length of $1024$ tokens. We train for up to $300$ iterations with early stopping on validation loss
for the token-matched setting, and extend this budget to $700$ iterations
with the same early-stopping criterion for the proportionate upsampling
mixture to accommodate its substantially larger training corpus.
For dataset partitioning, we apply distinct split ratios tailored to the scale of each corpus. For both the ShareGPT and Switchboard datasets, we partition the data into training, validation, and test subsets at a dialogue ratio of 7:1:2. Because the Fisher corpus is substantially larger, we adjust its dialogue ratio to 7.75:0.25:2 to prevent the validation set from becoming unnecessarily large for frequent validation.

\subsection{Preprocessing and Transformation Thresholds}
\label{sec:appendix_preprocess}

Several thresholds govern data preprocessing and the event-guided transformation pipeline. For HH dialogue filtering, we discard dialogues whose Word Error Rate exceeds $0.3$, removing both broken audio and cross-channel leakage where one speaker's audio bleeds into the other channel. For the event-guided transformation, the IPU pause threshold is set to $32$ milliseconds, each silence token represents $0.64$ seconds of silence, and the maximum utterance duration for heuristic backchannel identification is capped at $1$ second.

\section{Additional Experimental Results}
\label{sec:appendix_results}

\begin{figure*}[t]
  \includegraphics[width=0.45\linewidth]{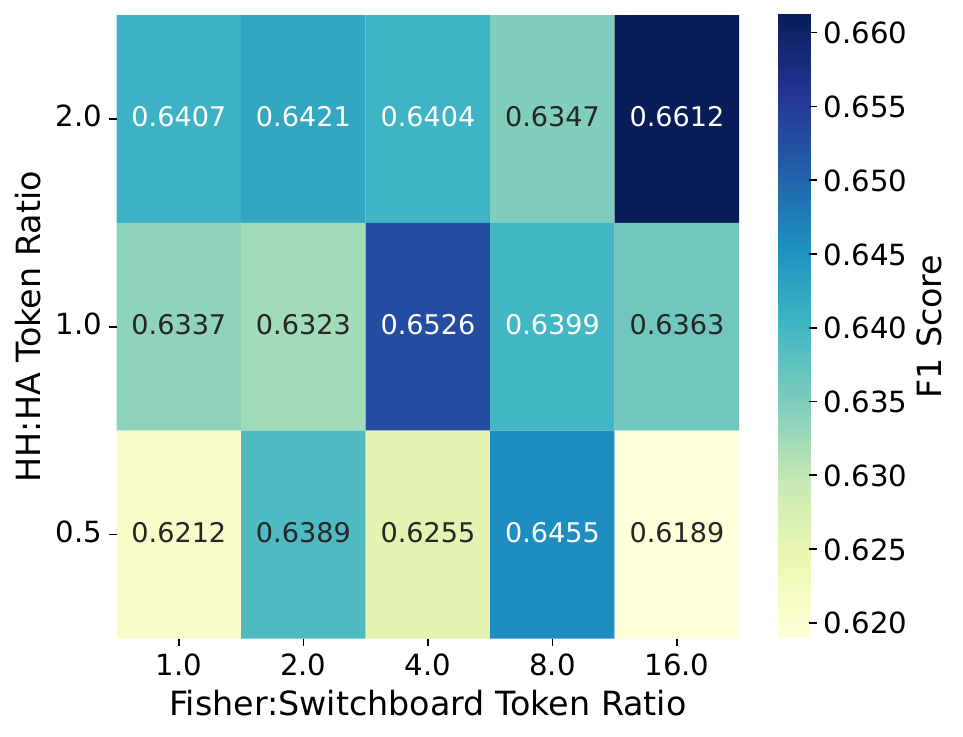} \hfill
  \includegraphics[width=0.45\linewidth]{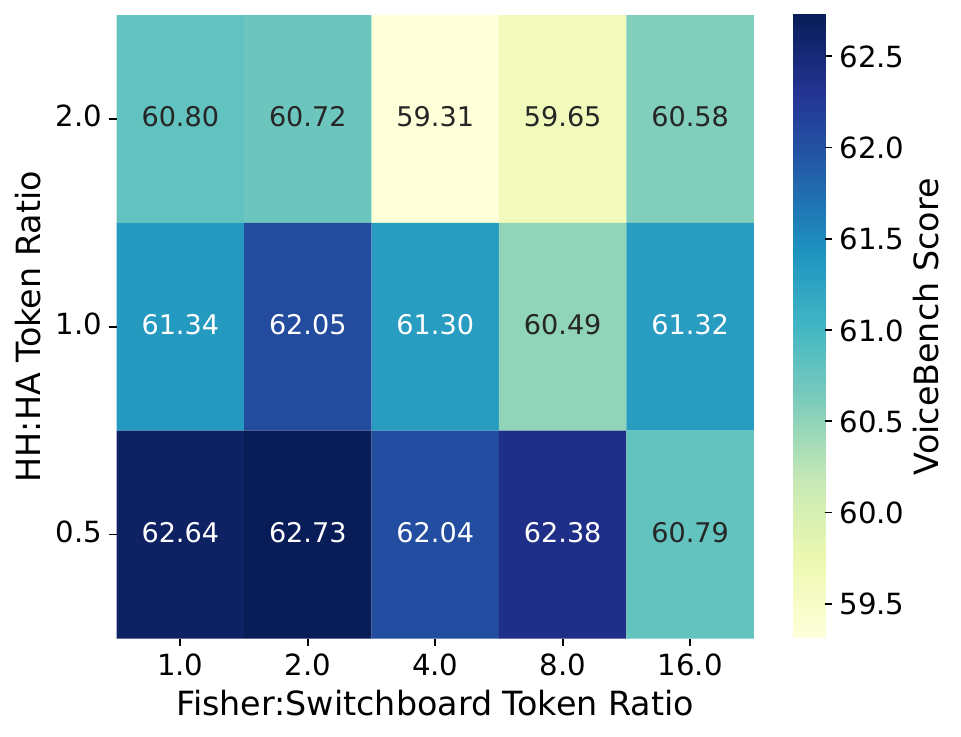}
  \caption{Impact of dataset mixture ratios on model performance. Left: turn-taking proficiency measured by F1 score. Right: semantic capability measured by VoiceBench.}
  \label{fig:ratio}
\end{figure*}

\begin{figure}[t]
  \centering
  \includegraphics[width=0.9\linewidth]{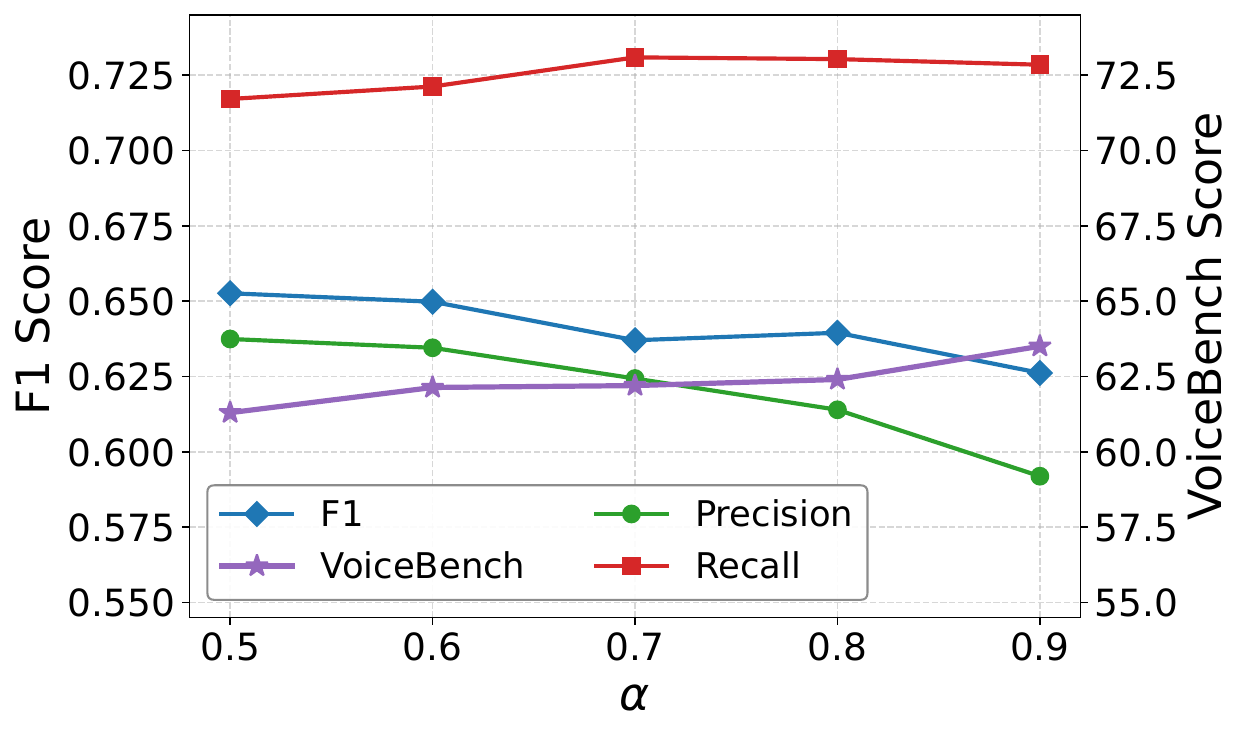}
  \caption{Impact of $\alpha$ in the SAC loss.}
  \vspace{-3mm}
  \label{fig:alpha}
\end{figure}

\begin{figure}[t]
  \centering
  \includegraphics[width=0.9\linewidth]{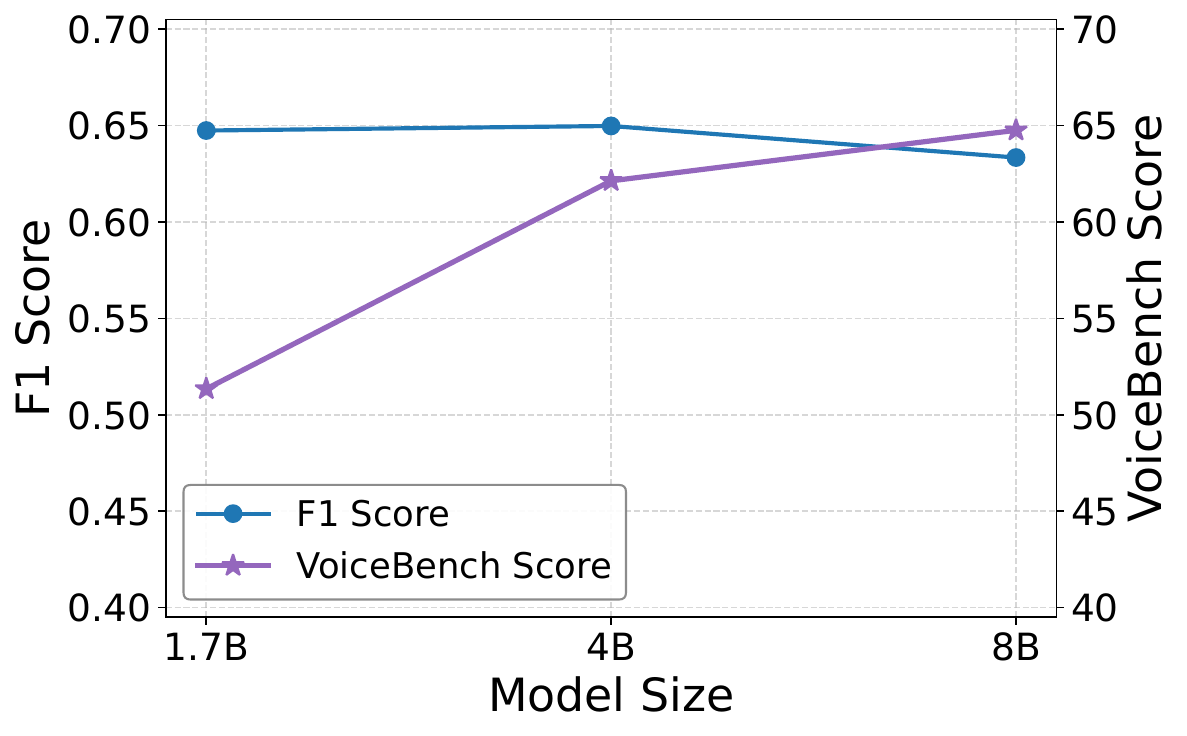}
  \caption{Impact of base model size.}
  \vspace{-3mm}
  \label{fig:model_size}
\end{figure}

\subsection{Analysis on Data Mixture, Loss, and Model Scale}
\label{sec:appendix_hyper_scaling}

This subsection reports analysis along three design axes: the mixture ratios of training data, the SAC loss hyperparameter $\alpha$, and the size of the base model. All experiments use the Faster-Whisper-based configuration and strictly constrain the total training token volume to match the NFSM baseline.

\paragraph{Data Mixture Ratios.}
Figure~\ref{fig:ratio} reports the effect of dataset mixture ratios under $\alpha=0.5$. The results demonstrate a clear trade-off: increasing the proportion of HH data consistently enhances turn-taking proficiency, whereas increasing the proportion of HA data improves semantic capability. To avoid biasing the model toward either capability, we adopt a 1:1 HH:HA ratio in all main experiments. Adjusting the internal Fisher-to-Switchboard ratio shows no significant trend, so we empirically set it to $4{:}1$.

\textbf{Impact of $\alpha$ in SAC Loss.}
Figure~\ref{fig:alpha} reports performance across different values of $\alpha$. As $\alpha$ increases, the VoiceBench score improves while the turn-taking F1 score decreases. To diagnose the F1 decline, we additionally plot the average recall and precision of state transition tokens on HH data. Interestingly, as $\alpha$ increases, recall actually rises and stabilizes, whereas precision drops significantly. This indicates that the SAC loss inherently aligns better with recall. This behavior is consistent with prior observations that standard language model pre-training via maximum likelihood estimation inherently favors recall over precision, assigning excessive probability mass to an unreliable tail of tokens \cite{Holtzman2020The, 10.5555/3540261.3540629}. Based on this analysis, we adopt $\alpha=0.6$ as the default in our main experiments to balance semantic preservation against turn-taking precision.

\paragraph{Impact of Base Model Size.}
Fixing $\alpha=0.6$, we further investigate the effect of scaling the base model, as shown in Figure~\ref{fig:model_size}. The VoiceBench score consistently improves with larger models, but the turn-taking F1 score peaks at the 4B scale. We hypothesize that larger language models may require substantially more effort to adequately fit the highly specialized pattern of the FSM tape.

\begin{table}[t]
\centering
\small
\setlength{\tabcolsep}{8pt}
\begin{tabular}{@{}lrr@{}}
\toprule
\textbf{Configuration} & \textbf{F1} $\uparrow$ & \textbf{VB} $\uparrow$ \\ \midrule
Ours & 0.6404 & 54.80 \\ \midrule
\multicolumn{3}{@{}l}{\textit{Ablation on Loss Function}} \\
\quad w/o SAC loss & 0.6234 & 54.49 \\ \midrule
\multicolumn{3}{@{}l}{\textit{Ablation on Training Data (w/o SAC loss)}} \\
\quad HH only        & 0.6376 & 19.71 \\
\quad HA only        & 0.3886 & 54.55 \\
\quad Synthetic only (NFSM) & 0.3025 & 44.24 \\ \bottomrule
\end{tabular}
\caption{Ablation study on the SAC loss and training data composition under SimulStreaming perception.}
\label{tab:ablation_simul}
\end{table}

\subsection{Ablation Study under SimulStreaming Perception}
\label{sec:appendix_simul_ablation}

To complement the ablation study in the main text, Table~\ref{tab:ablation_simul} reports the corresponding results under SimulStreaming perception. The trends are highly consistent with those of the Faster-Whisper-based model: the SAC loss effectively preserves turn-taking performance, and neither HH nor HA data alone is sufficient to jointly optimize semantic and turn-taking capabilities.

\begin{table}[t]
\centering
\small
\begin{tabular}{@{}llrr@{}}
\toprule
\textbf{Model} & \textbf{Perception} & \textbf{MiU F1} $\uparrow$ & \textbf{UiM PRR} $\uparrow$\\ \midrule
NFSM & --- & 0.7203 & 0.7260 \\ \midrule
\multirow{2}{*}{Ours} & Faster-Whisper & 0.7698 & 0.8037 \\
 & SimulStreaming & \textbf{0.7786} & \textbf{0.8367} \\ \bottomrule
\end{tabular}
\caption{Experiments of bidirectional interruption.}
\label{tab:interruption}
\end{table}

\begin{table*}[t]
\centering
\small
\setlength{\tabcolsep}{4pt}
\begin{tabular}{@{}lcccccccccc@{}}
\toprule
\textbf{Dimension} & \multicolumn{2}{c}{\textbf{Pause Handling}} & \multicolumn{3}{c}{\textbf{Backchannel}} & \multicolumn{2}{c}{\textbf{Smooth Turn Taking}} & \multicolumn{3}{c}{\textbf{User Interruption}} \\
\cmidrule(lr){2-3} \cmidrule(lr){4-6} \cmidrule(lr){7-8} \cmidrule(lr){9-11}
\textbf{Data} & Synthetic & Candor & \multicolumn{3}{c}{ICC} & \multicolumn{2}{c}{Candor} & \multicolumn{3}{c}{Synthetic} \\
\textbf{Metric} & TOR $\downarrow$ & TOR $\downarrow$ & TOR $\downarrow$ & Freq $\uparrow$ & JSD $\downarrow$ & TOR $\uparrow$ & Latency $\downarrow$ & TOR $\uparrow$ & GPT-4o $\uparrow$ & Latency $\downarrow$ \\
\midrule
dGSLM       & 0.934 & 0.935 & 0.691 & 0.015 & 0.934 & \textbf{0.975} & 0.352 & 0.917 & 0.201 & 2.531 \\
Moshi       & 0.985 & 0.980 & 1.000 & 0.001 & 0.957 & 0.941 & 0.265 & \textbf{1.000} & 0.765 & \textbf{0.257} \\
Freeze-Omni & 0.642 & 0.481 & 0.636 & 0.001 & 0.997 & 0.336 & 0.953 & 0.867 & \textbf{3.615} & 1.409 \\
Gemini Live & \textbf{0.255} & \textbf{0.310} & \textbf{0.091} & 0.012 & 0.896 & 0.655 & 1.301 & 0.891 & 3.376 & 1.183 \\
\midrule
Ours & 0.825 & 0.653 & 0.127 & \textbf{0.119} & \textbf{0.764} & 0.891 & \textbf{0.181} & 0.875 & 3.589 & 0.579 \\
\bottomrule
\end{tabular}
\caption{Complete results on Full-Duplex-Bench v1.0 across different conversational dimensions, where latency is presented in seconds.}
\label{tab:fdb_v1_full}
\end{table*}

\begin{table*}[t]
\centering
\small
\begin{tabular}{@{}llcccccc@{}}
\toprule
\textbf{Scenario} & \textbf{Class / Metric} & \textbf{Freeze-Omni} & \textbf{Moshi} & \textbf{Gemini} & \textbf{Sonic} & \textbf{GPT-4o} & \textbf{Ours} \\
\midrule
\multirow{6}{*}{USER\_INTR}
 & \textbf{RESPOND} $\uparrow$     & 0.72 & 0.50 & 0.33 & 0.24 & \textbf{0.78} & 0.69 \\
 & RESUME $\downarrow$    & 0.12 & 0.26 & 0.55 & 0.71 & \textbf{0.10} & 0.17 \\
 & UNCERTAIN $\downarrow$ & 0.03 & \textbf{0.00} & 0.01 & 0.01 & 0.02 & 0.05 \\
 & UNKNOWN $\downarrow$   & 0.13 & 0.25 & 0.10 & \textbf{0.04} & 0.12 & 0.09 \\
 \cmidrule(lr){2-8}
 & STOP (s) $\downarrow$  & 1.42 & 1.16 & 2.20 & 2.25 & \textbf{0.23} & 1.60 \\
 & RESP (s) $\downarrow$  & 1.35 & 1.47 & 2.62 & 2.75 & 1.50 & \textbf{1.12} \\
\midrule
\multirow{6}{*}{USER\_BACKCH}
 & RESPOND $\downarrow$   & 0.07 & 0.02 & 0.01 & \textbf{0.00} & 0.03 & \textbf{0.00} \\
 & \textbf{RESUME} $\uparrow$      & 0.80 & 0.06 & 0.93 & \textbf{0.98} & 0.70 & \textbf{0.98} \\
 & UNCERTAIN $\downarrow$ & 0.02 & \textbf{0.00} & 0.02 & \textbf{0.00} & 0.01 & \textbf{0.00} \\
 & UNKNOWN $\downarrow$   & 0.11 & 0.92 & 0.04 & \textbf{0.02} & 0.25 & \textbf{0.02} \\
 \cmidrule(lr){2-8}
 & STOP (s) $\uparrow$    & 0.66 & 0.42 & 0.66 & 0.64 & 0.21 & \textbf{0.69} \\
 & RESP (s) $\downarrow$  & 2.16 & 3.00 & 2.45 & 1.45 & \textbf{1.32} & 1.99 \\
\midrule
\multirow{6}{*}{TALKING\_OTHER}
 & RESPOND $\downarrow$   & 0.58 & 0.20 & \textbf{0.00} & 0.10 & 0.91 & 0.40 \\
 & \textbf{RESUME} $\uparrow$      & 0.25 & 0.19 & \textbf{0.99} & 0.90 & 0.02 & 0.44 \\
 & UNCERTAIN $\uparrow$   & 0.00 & 0.02 & 0.00 & 0.00 & 0.01 & \textbf{0.06} \\
 & UNKNOWN $\downarrow$   & 0.15 & 0.59 & 0.01 & \textbf{0.00} & 0.06 & 0.10 \\
 \cmidrule(lr){2-8}
 & STOP (s) $\uparrow$    & 1.39 & 0.87 & 1.69 & \textbf{1.77} & 0.18 & 1.35 \\
 & RESP (s) $\downarrow$  & 1.32 & 2.38 & 1.78 & 2.04 & \textbf{1.16} & 1.29 \\
\midrule
\multirow{6}{*}{BKG\_SPEECH}
 & RESPOND $\downarrow$   & 0.63 & 0.21 & 0.70 & \textbf{0.01} & 0.93 & 0.44 \\
 & \textbf{RESUME} $\uparrow$      & 0.25 & 0.07 & 0.30 & \textbf{0.98} & 0.04 & 0.36 \\
 & UNCERTAIN $\uparrow$   & 0.01 & 0.01 & 0.00 & 0.00 & 0.00 & \textbf{0.13} \\
 & UNKNOWN $\downarrow$   & 0.11 & 0.71 & \textbf{0.00} & 0.01 & 0.03 & 0.07 \\
 \cmidrule(lr){2-8}
 & STOP (s) $\uparrow$    & 0.98 & 0.54 & 0.95 & 1.05 & 0.18 & \textbf{1.16} \\
 & RESP (s) $\downarrow$  & 1.60 & 1.62 & 2.38 & 2.76 & 1.26 & \textbf{1.20} \\
\bottomrule
\end{tabular}
\caption{Complete results on Full-Duplex-Bench v1.5. Behavioral response distribution across the four overlap scenarios, with average stop and response latencies (s). Bold Class/Metric are the desired behavior for each scenario.}
\label{tab:fdb_v15_full}
\end{table*}

\subsection{Bidirectional Interruption Evaluation}
\label{sec:appendix_interruption}

We further evaluate the system's interruption handling capability by strictly replicating the protocol of NFSM.

For the Machine-interrupts-User (MiU) setting, we use GPT-4-Turbo to synthesize 600 dialogues in which the user's final statement is injected with a deliberate commonsense error. We decode the FSM step by step through this final statement and extract the response generated at the precise location where the model first proactively emits \verb|[S.SPEAK]| instead of \verb|[C.LISTEN]|. State transition tokens are then stripped to reconstruct a standard dialogue. To prevent egocentric bias \cite{koo-etal-2024-benchmarking}, we use Qwen3.5-27B as the evaluator, strictly following the assessment prompts of the original NFSM paper to compute the Proper Interruption Rate, the position-specific metrics $\text{ir}_\text{mid}$, $\text{ir}_\text{end}$, and MIR, and finally an aggregate F1 score.

For the User-interrupts-Machine (UiM) setting, we use GPT-4-Turbo to generate 600 dialogues evenly distributed across four interruption categories: denial, affirmation, environmental noises, and topic shifting. We insert \verb|[S.SPEAK]| at the end of the user's interruption to elicit an immediate FSM response. After removing state transition tokens from the dialogue, the average Proper Response Rate (PRR) is evaluated by the same Qwen3.5-27B judge using the original evaluation prompts.

As shown in Table~\ref{tab:interruption}, both perception-module variants of our model achieve substantially higher MiU F1 and UiM PRR scores than the NFSM baseline, demonstrating improvements in both turn-taking dynamics and semantic appropriateness within the specific sub-scenario of interruptions.

\subsection{Full Results on Full-Duplex-Bench}
\label{sec:appendix_fdb}

We report the complete Full-Duplex-Bench results underlying the merged Table~\ref{tab:fdb_merged} in the main text. Results of v1.0 are shown in Table~\ref{tab:fdb_v1_full}, while results of v1.5 are shown in Table~\ref{tab:fdb_v15_full}. The two versions probe complementary capabilities. v1.0 evaluates turn-taking over the course of a dialogue, covering whether the model holds back at intra-turn pauses, backchannels at human-like moments, takes the floor promptly at genuine turn boundaries, and reacts to interruptions. v1.5 instead isolates overlap handling while the model is speaking, testing whether it distinguishes overlapping speech that warrants yielding the floor from speech that does not. Both evaluate our Faster-Whisper-based proportionate upsampling model under the official protocols. All other numbers are quoted from the official report. Specially, the low Smooth Turn Taking latency is partly affected by the official metric, which by design clamps to zero any onset preceding the annotated turn boundary (e.g., backchannel).

\section{End-to-End Latency Breakdown}
\label{sec:appendix_latency}

\begin{table}[t]
\centering
\small
\begin{tabular}{@{}lrrr@{}}
\toprule
\textbf{Component} & \textbf{Mean (s)} & \textbf{P50 (s)} & \textbf{P90 (s)} \\ \midrule
Perception & 0.1689 & 0.1226 & 0.3347 \\
Cognitive  & 0.2786 & 0.2669 & 0.4357 \\
Motor      & 0.1481 & 0.1371 & 0.2024 \\ \midrule
Total      & 0.5955 & 0.5312 & 0.9105 \\ \bottomrule
\end{tabular}
\caption{End-to-end FTED latency of the FSM system under Faster-Whisper perception, broken down by module.}
\label{tab:latency_faster}
\end{table}


Table~\ref{tab:latency_faster} reports the end-to-end inference latency of the Faster-Whisper-based FSM system, measured by FTED. Latency is measured on a test set of 100 questions, constructed by randomly sampling 25 questions from each of the four VoiceBench subsets used in our semantic evaluation. The breakdown across the perception, cognitive, and motor modules demonstrates that the system maintains real-time responsiveness.

\section{Design of FSM}
\label{sec:appendix_fsm_design}

We build the FSM from scratch and introduce several techniques to improve its performance and robustness.

\textbf{Decoupled Temperatures for Transition and Response Tokens.} Although emitted from a shared autoregressive stream, state-transition and response tokens serve distinct purposes, necessitating decoupled decoding strategies. Transition tokens act as discrete control signals where sampling noise causes turn-taking failures; hence, we decode them greedily. Conversely, response tokens require standard sampling temperatures to maintain lexical diversity and generation quality. Operationally, each generation step initiates with a greedy single-token probe. If a transition token is emitted, the step commits it and terminates. Otherwise, this token serves as the clause prefix, and the remainder is sampled at the response temperature. Transition tokens proposed mid-clause by the sampler are discarded, deferring the transition decision to the next probe.

\textbf{Constrained Decoding over the Transition Vocabulary.} During decoding, we mask the logits of all state-transition tokens deemed illegal under the current FSM state. Although the fine-tuned model rarely produces illegal transitions, this hard masking provides deterministic guarantees against structural failures rather than relying on probabilistic safety.

\textbf{Bounded Look-ahead.} Generating the next clause only post-playback simplifies interruption handling but incurs latency gaps. Conversely, eager generation until turn completion eliminates gaps but wastes computation upon user intervention. We balance this trade-off via a bounded look-ahead policy: the controller generates one additional clause as long as fewer than $K$ clauses are \textit{in-flight} (i.e., dispatched to the TTS module but not yet fully played), idling once this bound is reached. 

\textbf{Interruption Handling.} Handling user interruptions gracefully presents a critical challenge. To support natural backchannels—where pausing before resumption sounds robotic—the agent continues speaking while the transition decision is computed. As a price, this incurs a temporal inconsistency that the tape before and after the decision diverge. In practice, we retrieve the currently playing clause from tape containing look-ahead tail and split it at the exact playback timestamp into a \textit{pre-clause} (already played) and a \textit{post-clause} (pending). We then execute one constrained decoding step using the pre-clause appended with the user utterance as the prefix. The resulting transition token dictates the tape rewriting strategy:
\begin{itemize}
    \item \verb|[C.SPEAK]|: The split clause is replaced by \textit{pre-clause + interruption +} \verb|[C.SPEAK]| \textit{+ post-clause}. The look-ahead tail is retained.
    \item \verb|[S.LISTEN.I]|: The split clause is replaced by \textit{pre-clause + interruption +} \verb|[S.LISTEN.I]|. TTS is truncated immediately, and the look-ahead tail is discarded.
    \item \verb|[S.LISTEN.N]|: The split clause is replaced by \textit{pre-clause + post-clause + interruption +} \verb|[S.LISTEN.N]|. The current clause is allowed to finish naturally, and the remaining look-ahead tail is discarded.
\end{itemize}

\textbf{Hallucination Filtering for the ASR Module.} Whisper-based ASR is prone to hallucinations, which can spuriously trigger the interruption mechanism. We mitigate this using a soft lexical filter: segments matching a predefined list of frequent hallucinations are discarded if their average token log-probability falls below $-0.3$. Otherwise, they are retained to preserve genuine user utterances.
\end{document}